\documentclass[a4paper,conference]{IEEEtran}
\IEEEoverridecommandlockouts

\usepackage[utf8]{inputenc} 
\usepackage[T1]{fontenc}
\usepackage[polutonikogreek,english]{babel}
\usepackage{cite}

\ifCLASSINFOpdf
  \usepackage[pdftex]{graphicx}
\else
\fi

\usepackage{times,amsmath,amssymb}

\usepackage{multirow}
\usepackage{array}
\usepackage[lofdepth,lotdepth]{subfig}
\usepackage{color}

\usepackage{algorithm2e}
\makeatletter
\renewcommand{\@algocf@capt@plain}{above}

\begin{document}


%
\title{Subspace Support Vector Data Description}

\author{\IEEEauthorblockN{Fahad Sohrab$^{\dag}$, Jenni Raitoharju$^{\dag}$, Moncef Gabbouj$^{\dag}$ and Alexandros Iosifidis$^{\ddag}$}
\IEEEauthorblockA{$^{\dag}$Laboratory of Signal Processing, Tampere University of Technology, Finland\\
$^{\ddag}$Department of Engineering, Electrical and Computer Engineering, Aarhus University, Denmark\\
{\{fahad.sohrab, jenni.raitoharju, moncef.gabbouj\}}@tut.fi, \:\:  alexandros.iosifidis@eng.au.dk}


}


%

\maketitle

\begin{abstract}
This paper proposes a novel method for solving one-class classification problems. The proposed approach, namely Subspace Support Vector Data Description, maps the data to a subspace that is optimized for one-class classification. In that feature space, the optimal hypersphere enclosing the target class is then determined. The method iteratively optimizes the data mapping along with data description in order to define a compact class representation in a low-dimensional feature space. We provide both linear and non-linear mappings for the proposed method. Experiments on 14 publicly available datasets indicate that the proposed Subspace Support Vector Data Description provides better performance compared to baselines and other recently proposed one-class classification methods.
\end{abstract}



%
\IEEEpeerreviewmaketitle


\IEEEpubidadjcol

\section{INTRODUCTION}
In data classification, the overall goal is to define a model that can classify data into a predefined set of classes. During the training phase, the parameters of the classification model are estimated using samples belonging to the classes of interest. When the classification task involves two or more classes, model training requires a sufficient number of samples from each class and the corresponding class labels. However, in cases where we are interested in the distinction of a class from all other classes, the application of multi-class classification methods is usually not appropriate.

In case where both the class of interest (here-after called positive class) and all other classes (used to form the negative class) are sufficiently represented in the training set, class-specific models can be employed \cite{liu2006class}\cite{iosifidis2017classspecific}\cite{iosifidis2015classspecific}\cite{iosifidis2016classspecific}, whereas, if the negative class is not sufficiently represented, one-class models should be applied. The main conceptual difference between class-specific and one-class models is that the former ones try to discriminate the positive class from every other class, while the latter ones try to describe the positive class without exploiting information related to negative samples. This is why one-class models can be applied in problems where only the positive class can be sufficiently sampled, while the negative one is either too rare or expensive to sample \cite{pimentel2014review}.

One-class classification problem has been tackled mainly by three approaches: density estimation, reconstruction and class boundary description \cite{martinus2001one}. For the density estimation approach, the Gaussian model, the mixture of Gaussians \cite{bishop1995neural} and the Parzen density \cite{parzen1962estimation} are the most popular ones \cite{ghasemigol2010new}. In reconstruction methods, the class is modelled by making some assumptions about the process which generates the target data. Some examples of reconstruction methods are based on K-means clustering, learning vector quantization and self-organizing maps \cite{kohonen1995learning}. In boundary description, a closed boundary around the target data is optimally formed. Support Vector Data Description (SVDD) \cite{tax2004support} is one of the popular boundary methods used for solving one-class classification problems, by defining a hypersphere enclosing the target class. The hypersphere of SVDD can be made more flexible using kernel methods \cite{tax2004support}. 

Other boundary methods have also been proposed for one-class classification. In \cite{scholkopf1999sv}, One-Class Support Vector Machine (OC-SVM) is proposed, in which the objective is to define the hyperplane that discriminates the data from the origin with maximum margin. It has also been proven that the solutions of SVDD and OC-SVM are equivalent for normalized data representations in the kernel space \cite{scholkopf2001learning} \cite{le2012unified}. In \cite{mygdalis2016graph}, Graph Embedded OC-SVM (GE-OC-SVM) and Graph Embedded SVDD (GE-SVDD) methods are introduced as extensions of \cite{scholkopf1999sv} and \cite{tax2004support}, respectively. These methods incorporate geometric class information expressed by generic graph structures in OC-SVM and SVDD optimization that acts as a regularizer to their solution. 

One-class classification has been used for many different applications. In \cite{razavi2017one}, one-class classification is used for detecting faults in induction motors. In \cite{sanchez2007one}, one-class classification, particularly SVDD, is used in remote sensing for mapping a specific land-cover class, illustrated with an example of classification of a local government district in Cambridgeshire, England. In \cite{sakla2011svdd}, an SVDD-based algorithm for target detection in hyperspectral images is developed. In \cite{popescu2009acoustic}, three different one-class classifiers, i.e., one-class Gaussian mixture, one-class SVM and one-class Nearest Neighbor are employed to label sound events as fall or part of the daily routine for elderly people based on sound signatures. In \cite{iosifidis2017oneClassVideoSummarization} one-class classification is used for video summarization based on human activities.

In this paper, we propose a novel method for generic one-class classification, namely Subspace Support Vector Data Description (S-SVDD). S-SVDD defines a model for the positive class in a low-dimensional feature space optimized for one-class classification. By allowing nonlinear data mappings, simple class models can be defined in the low-dimensional feature space that correspond to complex models in the original feature space. Such an approach allows us to simplify the information required for describing the class of interest, while at the same time it can provide a good performance in nonlinear problems. 

The rest of paper is organized as follows. The proposed S-SVDD method is described in detail in Section \ref{svddexplain}. Experiments conducted in order to evaluate its performance on generic one-class classification problems are provided in Section \ref{experiments}. Finally, conclusions are drawn in Section \ref{Conclusions}.

\section{SUBSPACE SUPPORT VECTOR DATA DESCRIPTION}\label{svddexplain}
Let us assume that the class to be modeled is represented by a set of vectors $\mathbf{x}_i, \: i=1,\dots,N$, living in a $D$-dimensional feature space (i.e. $\mathbf{x}_i \in \mathbb{R}^D$). Subspace Support Vector Data Description (S-SVDD) tries to determine a $d$-dimensional feature space ($d \le D$), in which the class can be optimally modeled. When linear projection is considered, the objective is to determine a matrix $\mathbf{Q} \in \mathbb{R}^{d \times D}$, such that:
\begin{equation}\label{eq:Y_i}
\mathbf{y}_i = \mathbf{Q} \mathbf{x}_i, \:\:, i=1,\dots,N,
\end{equation}
can be used in order to better model the class using a one-class classification model. We will describe how non-linear mappings can be exploited to this end using kernels in Subsection \ref{SS:KernelSSVDD}. 

The one-class classifier employed in this work is SVDD \cite{tax2004support}, which models the class by defining the hypesphere tightly enclosing the class. That is, given the data representation in the low-dimensional feature space $\mathbb{R}^d$, we want to determine the center of the class $\mathbf{a} \in \mathbb{R}^d$ and the corresponding radius $R$, by minimizing:
\begin{equation}\label{erfunc}
F(R,\textbf{a}) = R^2
\end{equation}  
such that all the training data are enclosed in the hypersphere, i.e.:
\begin{equation}\label{constraints0}
\| {\mathbf{Q}\mathbf{x}_i} - \mathbf{a} \|^2_{2} \le R^2, \:\: i=1,\dots,N.
\end{equation}


In order to define a tighter class boundary (and possibly handle the situation of outliers in the training data), a relaxed version of the above criterion is solved by introducing a set of slack variables $\xi_i$. That is, the optimization function to minimize becomes:
\begin{equation}\label{errorfunc}
F(R,\textbf{a}) = R^2 + C\sum_{i=1}^{N} \xi_i
\end{equation}  
under the constraints that most of the training data should lie inside the hyper-sphere, i.e.:
\begin{eqnarray}
\|\mathbf{Q}\mathbf{x}_i - \mathbf{a}\|_2^2 &\le& R^2 + \xi _i, \:\:i=1,\dots,N \label{constraints1}\\
\xi_i &\ge& 0, \:\:\:\:\:\:\:\:\:\:\:\:\:\:i=1,\dots,N. \label{constraints2}
\end{eqnarray}  
The parameter $C>0$ in (\ref{errorfunc}) is a regularization parameter which controls the trade-off between the volume of hypersphere and the training error caused by allowing outliers in the class description. $C$ is inversely proportional to the fraction of the expected outliers in the training set. Increasing the value of $C$ will allow more training samples to fall outside the class boundary.

The optimization problem in (\ref{errorfunc}), under the constraints in (\ref{constraints1}) and (\ref{constraints2}) corresponds to the original SVDD optimization problem optimized with respect to an additional parameter $\mathbf{Q}$ that is used to define the optimal data representations for one-class classification. In order to find the optimal parameter values, we apply Lagrange-based optimization. The Lagrangian function is given by:
\begin{eqnarray}\label{lang}
L(R,\mathbf{a},\alpha_i,\xi _i,\gamma_i,\mathbf{Q}) &=& {R^2} + C\sum_{i=1}^{N} \xi _i - \sum_{i=1}^{N} \gamma_i \xi_i \nonumber \\
&-& \sum_{i=1}^{N} \alpha _i \Big( R^2 + \xi _i - \mathbf{x}_i^T \mathbf{Q}^T \mathbf{Q} \mathbf{x}_i \nonumber \\
&+& 2\textbf{a}^T \textbf{Q} \textbf{x}_i- \textbf{a}^T\textbf{a} \Big)  
\end{eqnarray}
and should be maximized with respect to Lagrange multipliers $\alpha_i$ {$\geq 0, \gamma_i \geq 0$ }and minimized with respect to radius $R$, center $\textbf{a}$, slack variables ${\xi_i}$ and projection matrix $\mathbf{Q}$.

By setting the partial derivative to zero, we get:
\begin{eqnarray}
\frac{\partial L}{\partial R}=0 &\Rightarrow& \sum_{i=1}^{N} \alpha_i = 1 \label{der1} \\
\frac{\partial L}{\partial \mathbf{a}}=0 &\Rightarrow& \mathbf{a} = \sum_{i=1}^{N} \alpha_i \mathbf{Q}\mathbf{x}_i \label{der2} \\
\frac{\partial L}{\partial \xi _i}=0 &\Rightarrow& C- \alpha _i - \gamma _i  = 0 \label{der3} \\
\frac{\partial L}{\partial \mathbf{Q}}=0 &\Rightarrow& \mathbf{Q} = \left( \sum_{i=1}^N \alpha_i \mathbf{x}_i \mathbf{x}_i^T \right)^{-1} \left( \sum_{i=1}^N \alpha_i \mathbf{x}_i\mathbf{a}^T\right) \label{der4}
\end{eqnarray} 

From (\ref{der1})-(\ref{der4}), we can observe that the optimization parameters $\alpha_i$ and $\mathbf{Q}$ are inter-connected and, thus, they cannot be jointly optimized. In order to optimize (\ref{lang}) with respect to both $\alpha_i$ and $\mathbf{Q}$, we apply an iterative optimization process where, at each step, we fix one parameter and optimize the other, as will be described in the following subsections.

\subsection{Class description}\label{SS:ClassDescription}
Given a data projection matrix $\mathbf{Q}$, the data description step follows the standard SVDD-based solution. That is, substituting (\ref{eq:Y_i}), (\ref{der1}), (\ref{der2}) and (\ref{der3}) in (\ref{lang}) we obtain:
\begin{equation}\label{Lang2}
L = \sum_{i=1}^{N} \alpha _i \mathbf{y}_i^T  \mathbf{y}_i - \sum_{i=1}^{N}\sum_{j=1}^{N} \alpha _i \mathbf{y}_i^T \mathbf{y}_j \alpha _j.
\end{equation}  
Now, maximizing (\ref{Lang2}) gives the set of $\alpha_i,\:i=1,\dots,N$. The samples $\mathbf{y_i}=\mathbf{Q}\textbf{x}_i$ corresponding to values $\alpha{_i} > 0$ are the support vectors defining the data description. The samples $\mathbf{y_i}$ corresponding to values $0 < \alpha{_i} < C$ are on the boundary of the corresponding hypersphere, while those outside the boundary will correspond to values $\alpha_i = C$. For the samples $\mathbf{y_i}$ inside the boundary, the corresponding values of $\alpha_i$ will be equal to zero \cite{tax2004support}. Here we should note that whether a sample is a support vector or not, it is affected by the selection of the data projection matrix $\mathbf{Q}$, which is optimized based on the process described next.

\subsection{SVDD-based subspace learning}\label{SS:SubspaceLearning}
After determining the optimal set of $\alpha_i, \:i=1,\dots,N$, we optimize an augmented version of the Lagrangian function in (\ref{Lang2}):
\begin{equation}\label{lang3}
L= \sum_{i=1}^{N} \alpha _i  \mathbf{x}_i^T \mathbf{Q}^T \mathbf{Q} \mathbf{x}_i - \sum_{i=1}^{N}\sum_{j=1}^{N} \alpha_i \mathbf{x}_i^T \mathbf{Q}^T \mathbf{Q} \mathbf{x}_j \alpha_j + \beta\Psi,
\end{equation}  
where $\Psi$ is a regularization term expressing the class variance in the low-dimensional space having the form:
\begin{equation}
\label{generalconstraint}
\Psi = tr(\mathbf{Q}\mathbf{X} \lambda \lambda^T \mathbf{X}^T\mathbf{Q}^T).
\end{equation}  
$\beta$ is a regularization parameter controlling the importance of the regularization term in the update and $tr(\cdot)$ is the trace operator. We additionally impose the constraint $\mathbf{Q}\mathbf{Q}^T = \mathbf{I}$, in order to obtain a orthogonal projection. $\lambda \in \mathbb{R}^N$ is a vector controlling the contribution of each training sample in the regularization term and can take the following values:
\begin{enumerate}
    \item $\lambda_i = 0, \:i=1,\dots,N$: In this case the regularization term $\Psi$ becomes obsolete and $\mathbf{Q}$ is optimized using (\ref{Lang2}). This case is referred to as $\Psi_1$ here-after.
  
    \item $\lambda_i = 1, \:i=1,\dots,N$: In this case all training samples contribute to the regularization term $\Psi$ equally. That is, all samples are used in order to describe the variance of the class. This case is referred to as $\Psi_2$ here-after.
     
    \item $\lambda_i = \alpha_i, \:i=1,\dots,N$: In this case the samples belonging to the class boundary, as well as the outliers, are used to describe the class variance and regularize the update of $\mathbf{Q}$. This case is referred to as $\Psi_3$ here-after.

     \item $\lambda_i = \alpha_i^C, \:i=1,\dots,N$, where $\alpha^C \in \mathbb{R}^N$ is a vector with values $\alpha^C_i = \alpha_i$, if $\mathbf{Q}\mathbf{x}_i$ is a support vector, and $\alpha^C_i = 0$, otherwise. This case is referred to as $\Psi_4$ here-after.
\end{enumerate}

We update $Q$ by using the gradient of $L$, i.e.:
\begin{equation}\label{eqforQ1}
\Delta L = 2\sum_{i=1}^{N} \alpha_i \mathbf{Q} \mathbf{x}_i \mathbf{x}_i^T - 2\sum_{i=1}^{N}\sum_{j=1}^{N} \mathbf{Q} \mathbf{x}_i \mathbf{x}_j^T \alpha_i \alpha_j  + \beta\Delta \Psi,
\end{equation}  
where $\Delta \Psi$ is the derivative of (\ref{generalconstraint}) with respect to $\mathbf{Q}$, i.e.:
\begin{equation}\label{deltapsi1}
\Delta \Psi = 2\mathbf{Q} \mathbf{X} \lambda \lambda^T \mathbf{X}^T.
\end{equation}

\subsection{S-SVDD optimization}
In order to define both an optimized data projection matrix $\mathbf{Q}$ and the optimal data description in the resulting subspace, we iteratively apply the two processing steps described in subsections \ref{SS:ClassDescription} and \ref{SS:SubspaceLearning}, as described in Algorithm \ref{algo}. The $\alpha_i$'s computed by maximizing (\ref{Lang2}) are used in (\ref{eqforQ1}) to update $\mathbf{Q}$ through a gradient step using a learning rate parameter $\eta$. The projection matrix $\mathbf{Q}$ is orthogonalized and normalized in each iteration to force the orthogonality constraint before applying the data mapping.

\begin{algorithm}
  \caption{S-SVDD optimization}\label{algo}
\SetAlgoLined

\SetKwInOut{Input}{Input}
\SetKwInOut{Output}{Output}
\Input{$\mathbf{X}, \beta, \eta, d, C$}
 \Output{$\textbf{Q}$, $R$, $\mbox{\boldmath$\alpha$}$}  
 \vspace{3mm}
 // Initialize $\mathbf{Q}$\\
 Random initialization of $\mathbf{Q}$\;
 Orthogonalize $\mathbf{Q}$ using QR decomposition\;
 Row normalize $\mathbf{Q}$ using $l_2$ norm\;
 Initialize $k= 1$\;
 \vspace{3mm}
 \While{$k< k_{max}$ }{
    \vspace{3mm}
    // SVDD in the subspace defined by $\mathbf{Q}$ \\
    Calculate $\mathbf{Y}$ using (\ref{eq:Y_i})\;
    Calculate $\alpha_i,\:i=1,\dots,N$ using (\ref{Lang2})\;
    
    \vspace{3mm}
    // Update $\mathbf{Q}$ based on the SVDD solution\\
    Calculate $\Delta L$ using (\ref{generalconstraint})-(\ref{deltapsi1})\;
    Update $\mathbf{Q} \leftarrow \mathbf{Q} - \eta \Delta L$\;
    
    \vspace{3mm}
    // Normalize the updated $\mathbf{Q}$\\
    Orthogonalize $\mathbf{Q}$ using QR decomposition\;
    Row normalize $\mathbf{Q}$ using $l_2$ norm\;
   
      $k \leftarrow k+1$ 
   }
   \vspace{3mm}
   
   // SVDD in the optimized subspace\\
   Calculate $\mathbf{Y}$ using (\ref{eq:Y_i})\;
   Calculate $\alpha_i,\:i=1,\dots,N$ using (\ref{Lang2})\;
\end{algorithm}
\subsection{Non-linear data description}\label{SS:KernelSSVDD}
In order to exploit nonlinear mappings from $\mathbb{R}^D$ to $\mathbb{R}^d$ for one-class classification using the proposed S-SVDD, we follow the standard kernel-based learning approach \cite{scholkopf2001learning}. That is, the original data representations $\mathbf{x}_i \in \mathbb{R}^D, \:i=1,\dots,N$ are nonlinearly mapped to the so-called kernel space $\mathcal{F}$ using a nonlinear function $\phi(\cdot)$, such that $\mathbf{x}_i \in \mathbb{R}^D \rightarrow \phi(\mathbf{x}_i) \in \mathcal{F}$. In $\mathcal{F}$, a linear projection of all the training data to $\mathbb{R}^d$ is given by:
\begin{equation}\label{eq:Y_i_kernel}
\mathbf{y}_i = \mathbf{Q} \phi(\mathbf{x}_i), \:\:, i=1,\dots,N,
\end{equation}
where $\mathbf{Q} \in \mathbb{R}^{d \times |\mathcal{F}|}$ is a projection matrix of arbitrary dimensions \cite{scholkopf2001learning}. In order to calculate the data representations $\mathbf{y}_i, \:i=1,\dots,N$, we employ the kernel trick stating that $\mathbf{Q}$ can be expressed as a linear combination of the training data representations in $\mathcal{F}$ leading to:
\begin{equation}\label{eq:Y_i_kernel2}
\mathbf{y}_i = \mathbf{W} \mathbf{\Phi}^T \phi(\mathbf{x}_i) = \mathbf{W} \mathbf{k}_i, \:i=1,\dots,N,
\end{equation}
where $\mathbf{\Phi} \in \mathbb{R}^{|\mathcal{F}| \times N}$ is a matrix formed by the training data representations in $\mathcal{F}$, $\mathbf{W} \in \mathbb{R}^{d \times N}$ is a matrix containing the reconstruction weights of $\mathbf{W}$ with respect to $\mathbf{\Phi}$ and $\mathbf{k}_i$ is the $i$-th column of the so-called kernel matrix $\mathbf{K} \in \mathbb{R}^{N \times N}$ having elements equal to $\mathbf{K}_{ij} = \phi(\mathbf{x}_i)^T \phi(\mathbf{x}_j)$. In our experiments we use the RBF kernel, given by:
\begin{equation}\label{RBFkernel}
\mathbf{K}_{ij} = \exp  \left( \frac{ -\| \mathbf{x}_i - \mathbf{x}_j\|_2^2 }{ \sigma^2 } \right)
\end{equation}  
where $\sigma>0$ is a hyper-parameter scaling the Euclidean distance between $\mathbf{x}_i$ and $\mathbf{x}_j$.

In order to exploit the above-described nonlinear data mapping within the proposed S-SVDD method, we work as follows: for a given matrix $\mathbf{W}$, the training data $\mathbf{x}_i, \:i=1,\dots,N$ are mapped to $\mathbf{y}_i, \:i=1,\dots,N$ using (\ref{eq:Y_i_kernel2}) and $\alpha_i, \:i=1,\dots,N$ are calculated by optimizing (\ref{Lang2}). Subsequently, $\mathbf{W}$ is updated using:
\begin{equation}\label{eqforQ2}
\Delta L = 2\sum_{i=1}^{N} \alpha_i \mathbf{W} \mathbf{k}_i \mathbf{k}_i^T - 2\sum_{i=1}^{N}\sum_{j=1}^{N} \mathbf{W} \mathbf{k}_i \mathbf{k}_j^T \alpha_i \alpha_j  + \beta\Delta \Psi,
\end{equation}  
\begin{equation}\label{deltapsi2}
\Delta \Psi = 2\mathbf{W} \mathbf{K} \lambda \lambda^T \mathbf{K}^T.
\end{equation}

\subsection{Test phase}\label{SS:Test}
During testing, a sample $\mathbf{x}_{*} \in \mathbb{R}^D$ is mapped to its representation in the low-dimensional space $\mathbf{y}_{*} \in \mathbb{R}^d$ using (\ref{eq:Y_i}) (or (\ref{eq:Y_i_kernel2}) for the non-linear case) and its distance from the hypersphere center is calculated:
\begin{equation}\label{eqtest2}
\|\mathbf{y}_{*} - \mathbf{a}\|_2^2= \mathbf{y}_{*}^T\mathbf{y}_{*} - 2 \sum_{i=1}^{N} \alpha_i \mathbf{y}_{*}^T\mathbf{y}_i + \sum_{i=1}^{N}\sum_{j=1}^{N} \alpha_i \alpha_j \mathbf{y}_i^T\mathbf{y}_j.
\end{equation}
$\mathbf{y}_{*}$ is classified as positive when $\|\mathbf{y}_{*} - \mathbf{a}\|_2^2 \le R^2$ and as negative, otherwise.

\begin{table}[h]

    \caption{List of datasets used} 
  \begin{tabular}{ |p{0.3cm}|p{3cm}|p{0.9cm}|p{0.3cm}|p{2cm}|}

 \hline
No.& Dataset name&\textit{N}&\textit{D}&Target class\\ [0.3ex] 
 \hline\hline
 1&Balance scale&625&4&Left\\ 
 \hline
 2&Iris  & 150 & 4 &Iris-virginica\\
 \hline
 3&Lenses  & 24 & 4 &No contact lenses\\
 \hline
4& Seeds  & 210 & 7 &Kama\\
 \hline
5& Haberman's survival & 306 & 3 &Survived\\ 
  \hline
6& Qualitative bankruptcy&250 & 7&Bankrupt\\
  \hline
  7& User knowledge modeling&403 & 5 &Low\\
  \hline
   8&Pima Indians diabetes &768& 8 &No diabetes\\
  \hline
9&Banknote authentication  &1372&5 &No\\
  \hline
10& TA evaluation &151&5&High\\
  \hline

 11& PDelft pump  &1500& 64&Normal\\
  \hline
 12& Vehicle Opel & 864 &18&Opel\\ 
  \hline
  13& Sonar   & 208 &60&Mines\\
  \hline
 14&   Breast Wisconsin    & 699 & 9 &Malignant\\ 
  \hline

 \end{tabular}

 \label{datatable}
\end{table}

\begin{table*}
  \centering
   \caption{F1 measures on 14 datasets}
\begin{tabular}{lllllllllllllll}
\hline
\multicolumn{1}{|l|}{Dataset}       & \multicolumn{1}{l|}{1}              & \multicolumn{1}{l|}{2}              & \multicolumn{1}{l|}{3}              & \multicolumn{1}{l|}{4}              & \multicolumn{1}{l|}{5}              & \multicolumn{1}{l|}{6}              & \multicolumn{1}{l|}{7}              & \multicolumn{1}{l|}{8}              & \multicolumn{1}{l|}{9}              & \multicolumn{1}{l|}{10}             & \multicolumn{1}{l|}{11}             & \multicolumn{1}{l|}{12}             & \multicolumn{1}{l|}{13}             & \multicolumn{1}{l|}{14}             \\ \hline
\textbf{Linear}                      &                                     &                                     &                                     &                                     &                                     &                                     &                                     &                                     &                                     &                                     &                                     &                                     &                                     &                                     \\ \hline
\multicolumn{1}{|l|}{SVDD}           & \multicolumn{1}{l|}{0.703}          & \multicolumn{1}{l|}{0.762}          & \multicolumn{1}{l|}{0.609}          & \multicolumn{1}{l|}{0.774}          & \multicolumn{1}{l|}{0.834}          & \multicolumn{1}{l|}{0.686}          & \multicolumn{1}{l|}{0.634}          & \multicolumn{1}{l|}{0.791}          & \multicolumn{1}{l|}{0.764}          & \multicolumn{1}{l|}{0.485}          & \multicolumn{1}{l|}{0.846}          & \multicolumn{1}{l|}{0.853}          & \multicolumn{1}{l|}{0.625}          & \multicolumn{1}{l|}{0.958}          \\ \hline
\multicolumn{1}{|l|}{OC-SVM}          & \multicolumn{1}{l|}{0.688}          & \multicolumn{1}{l|}{0.612}          & \multicolumn{1}{l|}{0.394}          & \multicolumn{1}{l|}{0.619}          & \multicolumn{1}{l|}{0.644}          & \multicolumn{1}{l|}{0.562}          & \multicolumn{1}{l|}{0.532}          & \multicolumn{1}{l|}{0.529}          & \multicolumn{1}{l|}{0.657}          & \multicolumn{1}{l|}{\textbf{0.532}} & \multicolumn{1}{l|}{0.632}          & \multicolumn{1}{l|}{0.590}          & \multicolumn{1}{l|}{0.535}          & \multicolumn{1}{l|}{0.660}          \\ \hline
\multicolumn{1}{|l|}{S-SVDD $\Psi _1$}  & \multicolumn{1}{l|}{\textbf{0.907}} & \multicolumn{1}{l|}{\textbf{0.899}} & \multicolumn{1}{l|}{0.620}          & \multicolumn{1}{l|}{0.756}          & \multicolumn{1}{l|}{0.836}          & \multicolumn{1}{l|}{0.692}          & \multicolumn{1}{l|}{\textbf{0.960}} & \multicolumn{1}{l|}{0.786}          & \multicolumn{1}{l|}{\textbf{0.908}}          & \multicolumn{1}{l|}{0.482}          & \multicolumn{1}{l|}{0.856}          & \multicolumn{1}{l|}{\textbf{0.855}}          & \multicolumn{1}{l|}{0.618}          & \multicolumn{1}{l|}{0.957}          \\ \hline
\multicolumn{1}{|l|}{S-SVDD $\Psi _2$} & \multicolumn{1}{l|}{0.898}          & \multicolumn{1}{l|}{0.897}          & \multicolumn{1}{l|}{\textbf{0.724}}          & \multicolumn{1}{l|}{\textbf{0.827}}          & \multicolumn{1}{l|}{0.839}          & \multicolumn{1}{l|}{0.720}          & \multicolumn{1}{l|}{0.957}          & \multicolumn{1}{l|}{\textbf{0.793}} & \multicolumn{1}{l|}{0.889}          & \multicolumn{1}{l|}{0.502}          & \multicolumn{1}{l|}{0.857}          & \multicolumn{1}{l|}{\textbf{0.855}}          & \multicolumn{1}{l|}{0.599}          & \multicolumn{1}{l|}{\textbf{0.960}}          \\ \hline
\multicolumn{1}{|l|}{S-SVDD $\Psi _3$} & \multicolumn{1}{l|}{0.896}          & \multicolumn{1}{l|}{0.881}          & \multicolumn{1}{l|}{0.649}          & \multicolumn{1}{l|}{0.798}          & \multicolumn{1}{l|}{\textbf{0.841}}          & \multicolumn{1}{l|}{\textbf{0.722}}          & \multicolumn{1}{l|}{0.946}          & \multicolumn{1}{l|}{0.787}          & \multicolumn{1}{l|}{0.886}          & \multicolumn{1}{l|}{0.507}          & \multicolumn{1}{l|}{0.856}          & \multicolumn{1}{l|}{\textbf{0.855}}          & \multicolumn{1}{l|}{0.633}          & \multicolumn{1}{l|}{\textbf{0.960}}          \\ \hline
\multicolumn{1}{|l|}{S-SVDD $\Psi _4$} & \multicolumn{1}{l|}{0.896}          & \multicolumn{1}{l|}{0.868}          & \multicolumn{1}{l|}{0.694}          & \multicolumn{1}{l|}{0.778}          & \multicolumn{1}{l|}{0.821}          & \multicolumn{1}{l|}{0.715}          & \multicolumn{1}{l|}{0.954}          & \multicolumn{1}{l|}{0.784}          & \multicolumn{1}{l|}{0.852}          & \multicolumn{1}{l|}{0.458}          & \multicolumn{1}{l|}{\textbf{0.857}} & \multicolumn{1}{l|}{0.854}          & \multicolumn{1}{l|}{\textbf{0.638}} & \multicolumn{1}{l|}{0.953}          \\ \hline
\textbf{Non-linear}                      &                                     &                                     &                                     &                                     &                                     &                                     &                                     &                                     &                                     &                                     &                                     &                                     &                                     &                                     \\ \hline
\multicolumn{1}{|l|}{SVDD}           & \multicolumn{1}{l|}{0.734}          & \multicolumn{1}{l|}{0.827}          & \multicolumn{1}{l|}{0.413}          & \multicolumn{1}{l|}{\textbf{0.858}} & \multicolumn{1}{l|}{0.835}          & \multicolumn{1}{l|}{0.605}          & \multicolumn{1}{l|}{0.651}          & \multicolumn{1}{l|}{0.785}          & \multicolumn{1}{l|}{0.804}          & \multicolumn{1}{l|}{0.396}          & \multicolumn{1}{l|}{0.836}          & \multicolumn{1}{l|}{0.852}          & \multicolumn{1}{l|}{0.609}          & \multicolumn{1}{l|}{0.962}          \\ \hline

\multicolumn{1}{|l|}{OC-SVM}         & \multicolumn{1}{l|}{0.544}          & \multicolumn{1}{l|}{0.673}          & \multicolumn{1}{l|}{0.523}          & \multicolumn{1}{l|}{0.444}          & \multicolumn{1}{l|}{0.743}          & \multicolumn{1}{l|}{0.550}          & \multicolumn{1}{l|}{0.409}          & \multicolumn{1}{l|}{0.786}          & \multicolumn{1}{l|}{0.700}          & \multicolumn{1}{l|}{0.274}          & \multicolumn{1}{l|}{0.661}          & \multicolumn{1}{l|}{0.679}          & \multicolumn{1}{l|}{0.530}          & \multicolumn{1}{l|}{0.630}    

  \\ \hline
\multicolumn{1}{|l|}{GE-SVDD}        & \multicolumn{1}{l|}{0.757}          & \multicolumn{1}{l|}{0.857}          & \multicolumn{1}{l|}{0.314}          & \multicolumn{1}{l|}{0.799}          & \multicolumn{1}{l|}{0.811}          & \multicolumn{1}{l|}{0.554}          & \multicolumn{1}{l|}{0.654}          & \multicolumn{1}{l|}{\textbf{0.790}}          & \multicolumn{1}{l|}{0.797}          & \multicolumn{1}{l|}{0.484}          & \multicolumn{1}{l|}{0.830}          & \multicolumn{1}{l|}{0.847}          & \multicolumn{1}{l|}{0.550}          & \multicolumn{1}{l|}{\textbf{0.966}}               \\ \hline
\multicolumn{1}{|l|}{GE-OC-SVM}      & \multicolumn{1}{l|}{\textbf{0.815}}          & \multicolumn{1}{l|}{\textbf{0.869}}          & \multicolumn{1}{l|}{0.398}          & \multicolumn{1}{l|}{0.800}          & \multicolumn{1}{l|}{0.816}          & \multicolumn{1}{l|}{0.594}          & \multicolumn{1}{l|}{\textbf{0.658}}          & \multicolumn{1}{l|}{0.667}          & \multicolumn{1}{l|}{\textbf{0.930}} & \multicolumn{1}{l|}{\textbf{0.498}}          & \multicolumn{1}{l|}{0.613}          & \multicolumn{1}{l|}{0.788}          & \multicolumn{1}{l|}{0.593}          & \multicolumn{1}{l|}{0.962}          \\ \hline
\multicolumn{1}{|l|}{S-SVDD $\Psi _1$}  & \multicolumn{1}{l|}{0.635}          & \multicolumn{1}{l|}{0.725}          & \multicolumn{1}{l|}{\textbf{0.736}}          & \multicolumn{1}{l|}{0.727}          & \multicolumn{1}{l|}{0.842}          & \multicolumn{1}{l|}{0.700}          & \multicolumn{1}{l|}{0.518}          & \multicolumn{1}{l|}{0.786}          & \multicolumn{1}{l|}{0.728}          & \multicolumn{1}{l|}{0.472}          & \multicolumn{1}{l|}{0.836}          & \multicolumn{1}{l|}{\textbf{0.858}} & \multicolumn{1}{l|}{0.504}          & \multicolumn{1}{l|}{0.961}          \\ \hline
\multicolumn{1}{|l|}{S-SVDD $\Psi _2$} & \multicolumn{1}{l|}{0.662}          & \multicolumn{1}{l|}{0.573}          & \multicolumn{1}{l|}{0.603}          & \multicolumn{1}{l|}{0.540}          & \multicolumn{1}{l|}{\textbf{0.845}} & \multicolumn{1}{l|}{\textbf{0.762}} & \multicolumn{1}{l|}{0.523}          & \multicolumn{1}{l|}{\textbf{0.790}}          & \multicolumn{1}{l|}{0.717}          & \multicolumn{1}{l|}{0.473}          & \multicolumn{1}{l|}{\textbf{0.856}}          & \multicolumn{1}{l|}{\textbf{0.858}} & \multicolumn{1}{l|}{\textbf{0.637}}          & \multicolumn{1}{l|}{0.783}          \\ \hline
\multicolumn{1}{|l|}{S-SVDD $\Psi _3$} & \multicolumn{1}{l|}{0.734}          & \multicolumn{1}{l|}{0.694}          & \multicolumn{1}{l|}{0.624}          & \multicolumn{1}{l|}{0.719}          & \multicolumn{1}{l|}{0.838}          & \multicolumn{1}{l|}{0.620}          & \multicolumn{1}{l|}{0.578}          & \multicolumn{1}{l|}{0.785}          & \multicolumn{1}{l|}{0.720}          & \multicolumn{1}{l|}{0.417}          & \multicolumn{1}{l|}{\textbf{0.856}}          & \multicolumn{1}{l|}{\textbf{0.858}} & \multicolumn{1}{l|}{\textbf{0.637}}          & \multicolumn{1}{l|}{0.902}          \\ \hline
\multicolumn{1}{|l|}{S-SVDD $\Psi _4$} & \multicolumn{1}{l|}{0.495}          & \multicolumn{1}{l|}{0.700}          & \multicolumn{1}{l|}{\textbf{0.736}}          & \multicolumn{1}{l|}{0.774}          & \multicolumn{1}{l|}{0.841}          & \multicolumn{1}{l|}{0.632}          & \multicolumn{1}{l|}{0.562}          & \multicolumn{1}{l|}{0.572}          & \multicolumn{1}{l|}{0.703}          & \multicolumn{1}{l|}{0.474}          & \multicolumn{1}{l|}{0.832}          & \multicolumn{1}{l|}{\textbf{0.858}} & \multicolumn{1}{l|}{\textbf{0.637}}          & \multicolumn{1}{l|}{0.951}          \\ \hline
\end{tabular}

    \label{fm}
\end{table*}

 \begin{table*}
   \caption{Standard deviation of the F1 scores}
  \centering
\begin{tabular}{lllllllllllllll}
\hline
\multicolumn{1}{|l|}{Dataset} & \multicolumn{1}{l|}{1}     & \multicolumn{1}{l|}{2}     & \multicolumn{1}{l|}{3}     & \multicolumn{1}{l|}{4}     & \multicolumn{1}{l|}{5}     & \multicolumn{1}{l|}{6}     & \multicolumn{1}{l|}{7}     & \multicolumn{1}{l|}{8}     & \multicolumn{1}{l|}{9}     & \multicolumn{1}{l|}{10}    & \multicolumn{1}{l|}{11}    & \multicolumn{1}{l|}{12}    & \multicolumn{1}{l|}{13}    & \multicolumn{1}{l|}{14}    \\ \hline
\textbf{Linear}                         &                            &                            &                            &                            &                            &                            &                            &                            &                            &                            &                            &                            &                            &                            \\ \hline
\multicolumn{1}{|l|}{SVDD}    & \multicolumn{1}{l|}{0.014} & \multicolumn{1}{l|}{0.041} & \multicolumn{1}{l|}{0.152} & \multicolumn{1}{l|}{0.041} & \multicolumn{1}{l|}{0.009} & \multicolumn{1}{l|}{0.072} & \multicolumn{1}{l|}{0.032} & \multicolumn{1}{l|}{0.009} & \multicolumn{1}{l|}{0.010} & \multicolumn{1}{l|}{0.025} & \multicolumn{1}{l|}{0.005} & \multicolumn{1}{l|}{0.003} & \multicolumn{1}{l|}{0.033} & \multicolumn{1}{l|}{0.002} \\ \hline
\multicolumn{1}{|l|}{OC-SVM}   & \multicolumn{1}{l|}{0.074} & \multicolumn{1}{l|}{0.143} & \multicolumn{1}{l|}{0.257} & \multicolumn{1}{l|}{0.171} & \multicolumn{1}{l|}{0.055} & \multicolumn{1}{l|}{0.082} & \multicolumn{1}{l|}{0.071} & \multicolumn{1}{l|}{0.093} & \multicolumn{1}{l|}{0.015} & \multicolumn{1}{l|}{0.091} & \multicolumn{1}{l|}{0.027} & \multicolumn{1}{l|}{0.024} & \multicolumn{1}{l|}{0.083} & \multicolumn{1}{l|}{0.040} \\ \hline
\multicolumn{1}{|l|}{S-SVDD $\Psi _1$}  & \multicolumn{1}{l|}{0.022} & \multicolumn{1}{l|}{0.034} & \multicolumn{1}{l|}{0.154} & \multicolumn{1}{l|}{0.041} & \multicolumn{1}{l|}{0.007} & \multicolumn{1}{l|}{0.046} & \multicolumn{1}{l|}{0.017} & \multicolumn{1}{l|}{0.009} & \multicolumn{1}{l|}{0.026} & \multicolumn{1}{l|}{0.088} & \multicolumn{1}{l|}{0.002} & \multicolumn{1}{l|}{0.004} & \multicolumn{1}{l|}{0.022} & \multicolumn{1}{l|}{0.004} \\ \hline
\multicolumn{1}{|l|}{S-SVDD $\Psi _2$}  & \multicolumn{1}{l|}{0.026} & \multicolumn{1}{l|}{0.032} & \multicolumn{1}{l|}{0.136} & \multicolumn{1}{l|}{0.052} & \multicolumn{1}{l|}{0.017} & \multicolumn{1}{l|}{0.012} & \multicolumn{1}{l|}{0.019} & \multicolumn{1}{l|}{0.006} & \multicolumn{1}{l|}{0.031} & \multicolumn{1}{l|}{0.049} & \multicolumn{1}{l|}{0.001} & \multicolumn{1}{l|}{0.003} & \multicolumn{1}{l|}{0.058} & \multicolumn{1}{l|}{0.012} \\ \hline
\multicolumn{1}{|l|}{S-SVDD $\Psi _3$}  & \multicolumn{1}{l|}{0.029} & \multicolumn{1}{l|}{0.061} & \multicolumn{1}{l|}{0.157} & \multicolumn{1}{l|}{0.057} & \multicolumn{1}{l|}{0.009} & \multicolumn{1}{l|}{0.016} & \multicolumn{1}{l|}{0.016} & \multicolumn{1}{l|}{0.016} & \multicolumn{1}{l|}{0.039} & \multicolumn{1}{l|}{0.052} & \multicolumn{1}{l|}{0.001} & \multicolumn{1}{l|}{0.003} & \multicolumn{1}{l|}{0.048} & \multicolumn{1}{l|}{0.004} \\ \hline
\multicolumn{1}{|l|}{S-SVDD $\Psi _4$}  & \multicolumn{1}{l|}{0.024} & \multicolumn{1}{l|}{0.063} & \multicolumn{1}{l|}{0.118} & \multicolumn{1}{l|}{0.030} & \multicolumn{1}{l|}{0.038} & \multicolumn{1}{l|}{0.016} & \multicolumn{1}{l|}{0.031} & \multicolumn{1}{l|}{0.013} & \multicolumn{1}{l|}{0.110} & \multicolumn{1}{l|}{0.078} & \multicolumn{1}{l|}{0.001} & \multicolumn{1}{l|}{0.003} & \multicolumn{1}{l|}{0.022} & \multicolumn{1}{l|}{0.016} \\ \hline
\textbf{Non-linear}                     &                            &                            &                            &                            &                            &                            &                            &                            &                            &                            &                            &                            &                            &                            \\ \hline
\multicolumn{1}{|l|}{SVDD}    & \multicolumn{1}{l|}{0.020} & \multicolumn{1}{l|}{0.020} & \multicolumn{1}{l|}{0.276} & \multicolumn{1}{l|}{0.066} & \multicolumn{1}{l|}{0.011} & \multicolumn{1}{l|}{0.046} & \multicolumn{1}{l|}{0.027} & \multicolumn{1}{l|}{0.010} & \multicolumn{1}{l|}{0.011} & \multicolumn{1}{l|}{0.224} & \multicolumn{1}{l|}{0.008} & \multicolumn{1}{l|}{0.005} & \multicolumn{1}{l|}{0.042} & \multicolumn{1}{l|}{0.008} \\ \hline
\multicolumn{1}{|l|}{OC-SVM}   & \multicolumn{1}{l|}{0.164} & \multicolumn{1}{l|}{0.158} & \multicolumn{1}{l|}{0.331} & \multicolumn{1}{l|}{0.275} & \multicolumn{1}{l|}{0.139} & \multicolumn{1}{l|}{0.107} & \multicolumn{1}{l|}{0.233} & \multicolumn{1}{l|}{0.014} & \multicolumn{1}{l|}{0.073} & \multicolumn{1}{l|}{0.173} & \multicolumn{1}{l|}{0.114} & \multicolumn{1}{l|}{0.146} & \multicolumn{1}{l|}{0.075} & \multicolumn{1}{l|}{0.354} \\ \hline

\multicolumn{1}{|l|}{GE-SVDD} & \multicolumn{1}{l|}{0.029} & \multicolumn{1}{l|}{0.022} & \multicolumn{1}{l|}{0.312} & \multicolumn{1}{l|}{0.064} & \multicolumn{1}{l|}{0.045} & \multicolumn{1}{l|}{0.045} & \multicolumn{1}{l|}{0.052} & \multicolumn{1}{l|}{0.021} & \multicolumn{1}{l|}{0.023} & \multicolumn{1}{l|}{0.101} & \multicolumn{1}{l|}{0.007} & \multicolumn{1}{l|}{0.006} & \multicolumn{1}{l|}{0.042} & \multicolumn{1}{l|}{0.009} 
\\ \hline

\multicolumn{1}{|l|}{GE-OC-SVM} & \multicolumn{1}{l|}{0.039} & \multicolumn{1}{l|}{0.056} & \multicolumn{1}{l|}{0.368} & \multicolumn{1}{l|}{0.071} & \multicolumn{1}{l|}{0.026} & \multicolumn{1}{l|}{0.131} & \multicolumn{1}{l|}{0.058} & \multicolumn{1}{l|}{0.261} & \multicolumn{1}{l|}{0.019} & \multicolumn{1}{l|}{0.063} & \multicolumn{1}{l|}{0.188} & \multicolumn{1}{l|}{0.121} & \multicolumn{1}{l|}{0.090} & \multicolumn{1}{l|}{0.009} \\ \hline
\multicolumn{1}{|l|}{S-SVDD $\Psi _1$}  & \multicolumn{1}{l|}{0.006} & \multicolumn{1}{l|}{0.058} & \multicolumn{1}{l|}{0.060} & \multicolumn{1}{l|}{0.178} & \multicolumn{1}{l|}{0.010} & \multicolumn{1}{l|}{0.029} & \multicolumn{1}{l|}{0.036} & \multicolumn{1}{l|}{0.004} & \multicolumn{1}{l|}{0.039} & \multicolumn{1}{l|}{0.029} & \multicolumn{1}{l|}{0.047} & \multicolumn{1}{l|}{0.000} & \multicolumn{1}{l|}{0.282} & \multicolumn{1}{l|}{0.018} \\ \hline
\multicolumn{1}{|l|}{S-SVDD $\Psi _2$}  & \multicolumn{1}{l|}{0.053} & \multicolumn{1}{l|}{0.124} & \multicolumn{1}{l|}{0.340} & \multicolumn{1}{l|}{0.089} & \multicolumn{1}{l|}{0.004} & \multicolumn{1}{l|}{0.048} & \multicolumn{1}{l|}{0.051} & \multicolumn{1}{l|}{0.002} & \multicolumn{1}{l|}{0.012} & \multicolumn{1}{l|}{0.050} & \multicolumn{1}{l|}{0.002} & \multicolumn{1}{l|}{0.000} & \multicolumn{1}{l|}{0.000} & \multicolumn{1}{l|}{0.100} \\ \hline
\multicolumn{1}{|l|}{S-SVDD $\Psi _3$}  & \multicolumn{1}{l|}{0.013} & \multicolumn{1}{l|}{0.027} & \multicolumn{1}{l|}{0.353} & \multicolumn{1}{l|}{0.059} & \multicolumn{1}{l|}{0.008} & \multicolumn{1}{l|}{0.135} & \multicolumn{1}{l|}{0.074} & \multicolumn{1}{l|}{0.029} & \multicolumn{1}{l|}{0.017} & \multicolumn{1}{l|}{0.133} & \multicolumn{1}{l|}{0.002} & \multicolumn{1}{l|}{0.000} & \multicolumn{1}{l|}{0.000} & \multicolumn{1}{l|}{0.079} \\ \hline
\multicolumn{1}{|l|}{S-SVDD $\Psi _1$}  & \multicolumn{1}{l|}{0.369} & \multicolumn{1}{l|}{0.035} & \multicolumn{1}{l|}{0.060} & \multicolumn{1}{l|}{0.049} & \multicolumn{1}{l|}{0.007} & \multicolumn{1}{l|}{0.089} & \multicolumn{1}{l|}{0.058} & \multicolumn{1}{l|}{0.330} & \multicolumn{1}{l|}{0.021} & \multicolumn{1}{l|}{0.037} & \multicolumn{1}{l|}{0.057} & \multicolumn{1}{l|}{0.000} & \multicolumn{1}{l|}{0.000} & \multicolumn{1}{l|}{0.021} \\ \hline
\end{tabular}

   \label{Sensitivity}
\end{table*}

\begin{figure*}[ht]
	\centering
	\includegraphics[scale=0.16]{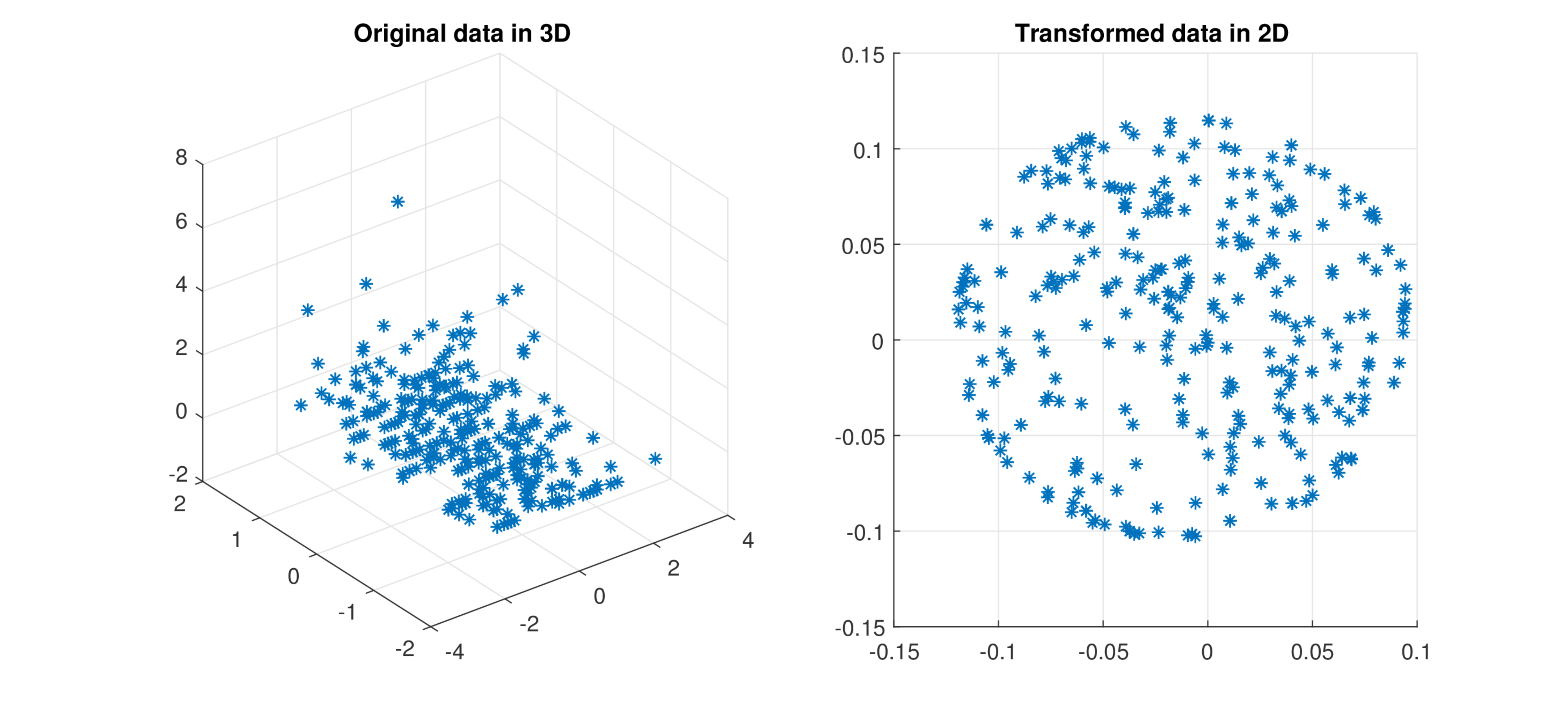}
	\caption{Transforming dataset 5 (Haberman's survival) from 3-dimensional feature space to 2-dimensional feature space using the proposed method (kernel S-SVDD $\Psi _1$)}
	\label{dimensionreduction}
\end{figure*}

\section{Experiments}\label{experiments}

\subsection{Data-sets, evaluation criteria and experimental setup}
We performed experiments on the datasets listed in Table \ref{datatable}. Datasets 1-10 are downloaded from UCI website \cite{Lichman:2013}, while datasets 11-14 are downloaded from TU delft pattern recognition lab website \cite{delftOCC}. The datasets with more than two classes were converted to a positive class and a negative class by considering the class with the  majority of samples as the positive class and others as the negative class. Table \ref{datatable} shows the target class of the each dataset in the last column. 

In binary classification, a machine learning model can make two kinds of errors during testing. It can either wrongly predict a data sample from the positive class as negative or a negative data sample as positive. In one-class classification, the focus is on the target class and usually it is of greater interest to predict the positive class accurately. Recall, also called sensitivity, hit rate, or true positive rate, is the proportion of correctly classified positive samples during the test:
\begin{equation}
\label{sensitivity}
Recall=\frac{tp}{p},
\end{equation}  
where $tp$ is the total number of correctly classified positive samples and $p$ is the total number of positive samples in the data. Recall is used to evaluate classification results in cases, where it is more important to predict the positive class accurately. Another metric used to evaluate machine learning algorithm is precision, which is the proportion of correctly classified samples among those classified into the positive class:
\begin{equation}
\label{precision}
Precision=\frac{tp}{tp+fp},
\end{equation}  
where $fp$ is an acronym for false positives, i.e., the number of samples incorrectly predicted as positive during the test. A perfect precision score of 1.0 means that every sample classified as positive is from the positive class. In other words, a low precision score indicates a large number of false positives. F1 measure takes into account both precision and recall. It is defined as their harmonic mean as
\begin{equation}
\label{f1score}
F1 = 2 * \frac{Precision * Recall} {Precision + Recall}.
\end{equation}  
We use (\ref{f1score}) for evaluating and comparing performance of the proposed algorithm with competing methods.

To perform our experiments we divided our datasets into train and test sets. We performed our experiments on each dataset by selecting 70 percent of the data for training and the remaining 30 percent for testing. The 70-30 train and test sets were selected randomly 5 times to check the performance of each model robustly. Thus, in total we created 5 train-test (70-30\%) partitions for each dataset. The proportion of each class in each set follows the original proportions. Also for the datasets having originally more than two classes, the positive and negative class labels were assigned after the subset for training and testing were created as described. 

We selected the parameters for the proposed method by 5-fold cross-validation over each training set according to the best average F1 measure and then used them to train the final model using the whole training set. Whenever we trained a model, only positive samples were used. We selected the value of the parameter $\beta$ as $10^l$, where $l=-4,\dots,4$, $\sigma$ is the scaled version of the mean distance between the training samples using a scaling factor $10^l$, where $l=-3,\dots,3$, and $C$ from $[0.01, 0.05, 0.1, 0.2, 0.3, 0.4, 0.5, 0.6]$. The subspace dimension $d$ for datasets having more than 10 dimensional feature space was restricted to a maximum of 10, i.e., $d=1,\dots,10$. For datasets with $D\leq10$, we set $d=1,\dots,D$.

We compared our results with the original SVDD (linear and kernel), OC-SVM (linear and kernel), GE-OC-SVM and GE-SVDD. The parameters were selected using a similar 5-fold cross-validation approach and the common parameters were selected from the ranges given above. Other parameters were selected as in the corresponding research papers.

\subsection{Experimental results}

Fig. \ref{dimensionreduction}, illustrates an example transformation of all the data samples of dataset 5 (Haberman’s survival) from the original 3-dimensional feature space to a lower 2-dimensional feature space using the non-linear version of the proposed S-SVDD method with the constraint $\Psi _1$ (see subsection \ref{SS:SubspaceLearning}). The figure shows the capability of the proposed method to transform the data to a compact form which is more suitable to be enclosed by a hypersphere.

In Tables \ref{fm} and \ref{Sensitivity}, we report the average F1 measure  and the standard deviation of F1 measure for the evaluated linear and non-linear methods. The linear version of the proposed S-SVDD clearly outperforms all other linear methods. Only for dataset 10, OC-SVM achieves a higher performance. The non-linear version of S-SVDD outperformed other non-linear methods on datasets 3, 5, 6, 11, 12 and 13. For dataset 8, GE-SVDD and S-SVDD $\Psi_2$ achieved the same results. GE-OC-SVM obtained the best results on datasets 1, 2, 7, 9 and 10.  Compared to the baseline methods (SVDD and OC-SVM), S-SVDD shows a clear improvement.

For datasets 12 and 13, the non-linear versions of S-SVDD  (except for $\Psi_1$ for dataset 13) have zero standard deviation. A closer inspection of the results shows that, in these cases, the obtained mapping and data description classify all the test samples as positive, due to the selection of small values for the hyper-parameter $C$ \cite{chang2013revisit}. A tighter fitted hypersphere on the training data may possibly lead to more meaningful results, which could be achieved by restricting the range of the $C$ values used during the cross-validation process applied on the training data for hyper-parameter selection of the proposed method. 

When comparing the different regularization terms $\Psi$ used with the proposed method, $\Psi_2$ achieves the best performance most often with both linear and non-linear versions. In $\Psi_2$, all training samples contribute to the regularization term equally.

\section{CONCLUSION}\label{Conclusions}
In this paper, we proposed a new method for one-class classification. The proposed S-SVDD method maps the original data to a lower dimensional feature space, which is more suitable for one-class classification. The method iteratively optimizes the mapping to the new subspace and the data description in that feature space. Both linear and non-linear versions were defined along with four different regularization terms. 

We performed experiments on 14 different publicly available datasets. Our experiments showed that the proposed method yields better results than the baselines and competing one-class classification methods in majority of the cases. A constraint that uses all samples for describing the data variance leads to the best results for S-SVDD. 

In the future, we intend to try the proposed S-SVDD method with different kernels and design new regularization terms. We will also evaluate a similar mapping approach in combination with other already established one-class classification methods.

\section*{Acknowledgement}
This work was supported by the NSF-TEKES Center for Visual and Decision Informatics project Co-Botics, jointly sponsored by Tieto Oy Finland and CA Technologies.



%

\bibliography{bibliography}
\bibliographystyle{IEEEtran}

\end{document}